\documentclass[11pt]{article}
     \usepackage[preprint]{neurips_2020}

\usepackage{natbib}
\setcitestyle{numbers,square}
\usepackage[utf8]{inputenc} 
\usepackage[T1]{fontenc}    
\usepackage{hyperref}       
\usepackage{url}            
\usepackage{booktabs}       
\usepackage{amsfonts}       
\usepackage{nicefrac}       
\usepackage{microtype}      

\usepackage{graphicx}    
\usepackage{amsmath}
\usepackage{amsfonts}
\usepackage{amssymb}
\usepackage{bbm}
\usepackage{algorithm}
\usepackage{algorithmic}

\title{A Simple Text to Video Model via Transformer}

\author{%
   Technical Report 
    \\
  Gang Chen \\
  \texttt{info@vividitytech.com} \\
}

\begin{document}         
\date{}

\maketitle
\begin{abstract}
We present a general and simple text to video model based on Transformer. Since both text and video are sequential data, we encode both texts and images into the same hidden space, which are further fed into Transformer to capture the temporal consistency and then decoder to generate either text or images. Considering the image signal may become weak in the long sequence, we introduce the U-Net to reconstruct image from its noised version. Specifically, we increase the noise level to the original image in the long sequence, then use the $down$ module from U-Net to encode noised images, which are further input to transformer to predict next clear images. We also add a constraint to promote motion between any generated image pair in the video. 
We use GPT2 and test our approach on UCF101 dataset and show it can generate promising videos. 

\end{abstract}

\section{Introduction}
Text to video has gained popularity in computer vision and machine learning community. At the very beginning, most work focus on how to generate images, such as GAN \cite{Goodfellow2014generative} and VAE \cite{Kingma2013} both of which have shown impressive image and speech synthesis results. Diffusion probabilistic models \cite{Sohldickstein2015deep,Ho2020denoising,Chen2022speed} have recently shown high quality image generations. Recently Ho et al. proposed video diffusion models \cite{Ho2022video}, which is a natural extension of the standard 2D image to 3D architecture. Imagen \cite{Saharia2022photorealistic} is a text-to-image diffusion model which is conditional on the text embedding from language models (e.g. T5). Imagen generates an unprecedented degree of photorealism given text prompt and boosts both sample fidelity and image-text alignment much more than increasing the size of the image diffusion model. Ho et al. extended Imagen and presented Imagen Video \cite{Ho2022imagen}, a text-conditional video generation system based on a cascade of video diffusion models. Given a text prompt, Imagen Video generates high definition videos using a base video generation model and a sequence of interleaved spatial and temporal video super-resolution models. However, Imagen Video requires videos with fixed length to construct 3D convolution process \cite{Ho2022video}.

VideoGPT \cite{Yan2019} is a conceptually simple architecture which uses VQ-VAE that learns downsampled discrete latent representations of a raw video by employing 3D convolutions and axial self-attention, and then leverage transformer to autoregressively model the discrete latents using spatio-temporal position encodings. Make-A-Video \cite{Singer2022makeavideo} is an approach for directly translating the tremendous recent progress in Text-to-Image (T2I) generation to Text-to-Video (T2V). Its intuition is simple: learn what the world looks like and how it is described from paired text-image data, and learn how the world moves from unsupervised video footage. 

Unfortunately these methods mentioned above either required the fixed video length for training or only generate the constrained videos with the same background scene. In this work, we present an approach to train on (text, video) pairs with varied lengths and different scenes based on transformer framework. In addition, to handle the possible weak signal in the long sequence, we introduce U-Net to reconstruct the video data.  

Specifically, we encode both text and video into the same hidden space, then we leverage transformer to capture the temporal and spatial consistency between video frames. In addition, we reconstruct the video data with noise to handle the long sequence scenario. For the text to video, we can either use a simple decoder or a conditional diffusion model to generate image, then capture the motion with temporal and spatial constrains.
One possible issue is the generated videos may concentrate on certain scene, so we add a constraint to promote motion. Considering the limited computation power, we use a simple decoder and train our model in a end-to-end fashion. Because there are limited text and video training dataset available at this moment, we focus on the UCF 101 action dataset. We select 60 categories of action from UCF 101 dataset, and label about 1 to 5 videos for each type of action with text descriptions. We tested our approach on this dataset, and show it can generate meaningful videos given a text prompt\footnote{Our implementation is available at https://github.com/vividitytech/text2videoGPT}.

\section{Model}
We present a simple text to video model via transformer. In the following parts, we will introduce the language models and then discuss how to combine transformer and U-Net to generate videos from texts.
\subsection{Background}
Given a vocabulary $\mathcal{V}$ and an ordered sequence of symbols (or tokens) $(x_1,x_2, ..., x_{n})$ with $x_i  \in \mathcal{V}$,  the language model \cite{Bengio2003} is defined as the joint probability over sequences of tokens $\mathcal{V}^n$, which is factorized into product of conditional probabilities
\begin{align}\label{eq:lm}
p(x_1,x_2, ..., x_{n}; \theta) = \prod_{1 \le i <n} p(x_i| x_1, x_2,..., x_{i-1}; \theta)
\end{align}
where the vocabulary $\mathcal{V}$ is a large but finite set, and $\theta$ is the model parameter. $p(x_i | x_1, x_2,..., x_{i-1})$ is conditional probability to predict next word given the previous sequences.

Many NLP problems can be formulated as $p(Y|X; \theta)$, where $X \in \mathcal{V}^n$ is the input sequence and $Y \in \mathcal{V}^m$ is the output.   
There have been many models that can compute these conditional probabilities, such as recurrent neural networks LSTM \cite{HochSchm97} and self-attention Transformer \cite{Vaswani2017attention}. Especially the transformer architecture have significant improvements in the expressiveness and accuracy of models \cite{Radford2018,Brown2020}. To learn the model parameters $\theta$, we can use cross entropy loss: 
\begin{align}\label{eq:loss1}
L_i(\theta)  = - \log p(y_i)  =  - \log p(y_i | y_{<i}, X; \theta) 
\end{align}
where only $y_i$ holds and other tokens in $\mathcal{V} \backslash y_i$ are zeros.
Then, the cross entropy error over the sequence of size $m$ is:
\begin{align}\label{eq:loss}
 L(\theta) = \sum_{i=1}^m L_i(\theta) = - \sum_{i=1}^m  \log p(y_i) 
\end{align}

While predicting the next symbol $\hat{y}_i \sim p(y_i | y_{<i}, X; \theta)$, we can either sample it or take a greedy strategy to select $\hat{y}_i$ with maximum probability. Note that the conditional probability $p(y_i) = p(y_i | y_{<i}, X; \theta)$ to predict next token is a discrete space $\mathcal{V}$, which is constrained by the vocabulary size. Compared to text, the image space is significant large and it is a much challenge problem to generate video from text.

\begin{figure}[t!]
\centering
\begin{tabular}{c}
\includegraphics[trim=5mm 35.0mm 5mm 18mm, clip,  width=12.2cm]{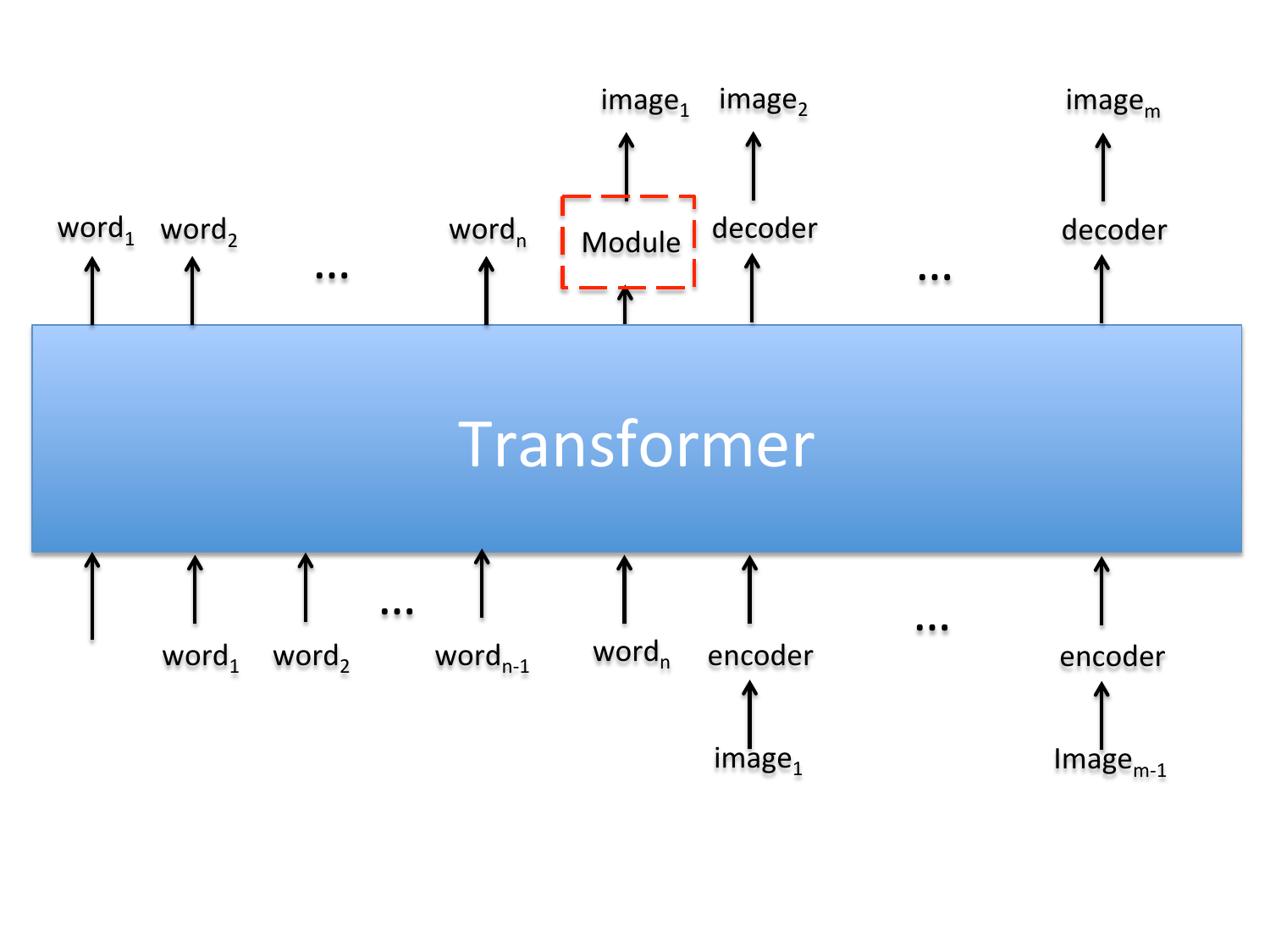}
\end{tabular}
\caption{The figure shows the architecture to encode both text and video using transformer, where the module inside red dash line can be any deep nets to generate image.}
\label{Fig:rlcomp}
\end{figure}

\subsection{The text to video model}
What if $Y=\{y_1, y_2,..., y_m \}$ is a video, not text? In this part, we will introduce how to extend Transformer to handle both texts and videos. 

\begin{figure}[t!]
\centering
\begin{tabular}{c}
\includegraphics[trim=5mm 35.0mm 5mm 18mm, clip,  width=12.2cm]{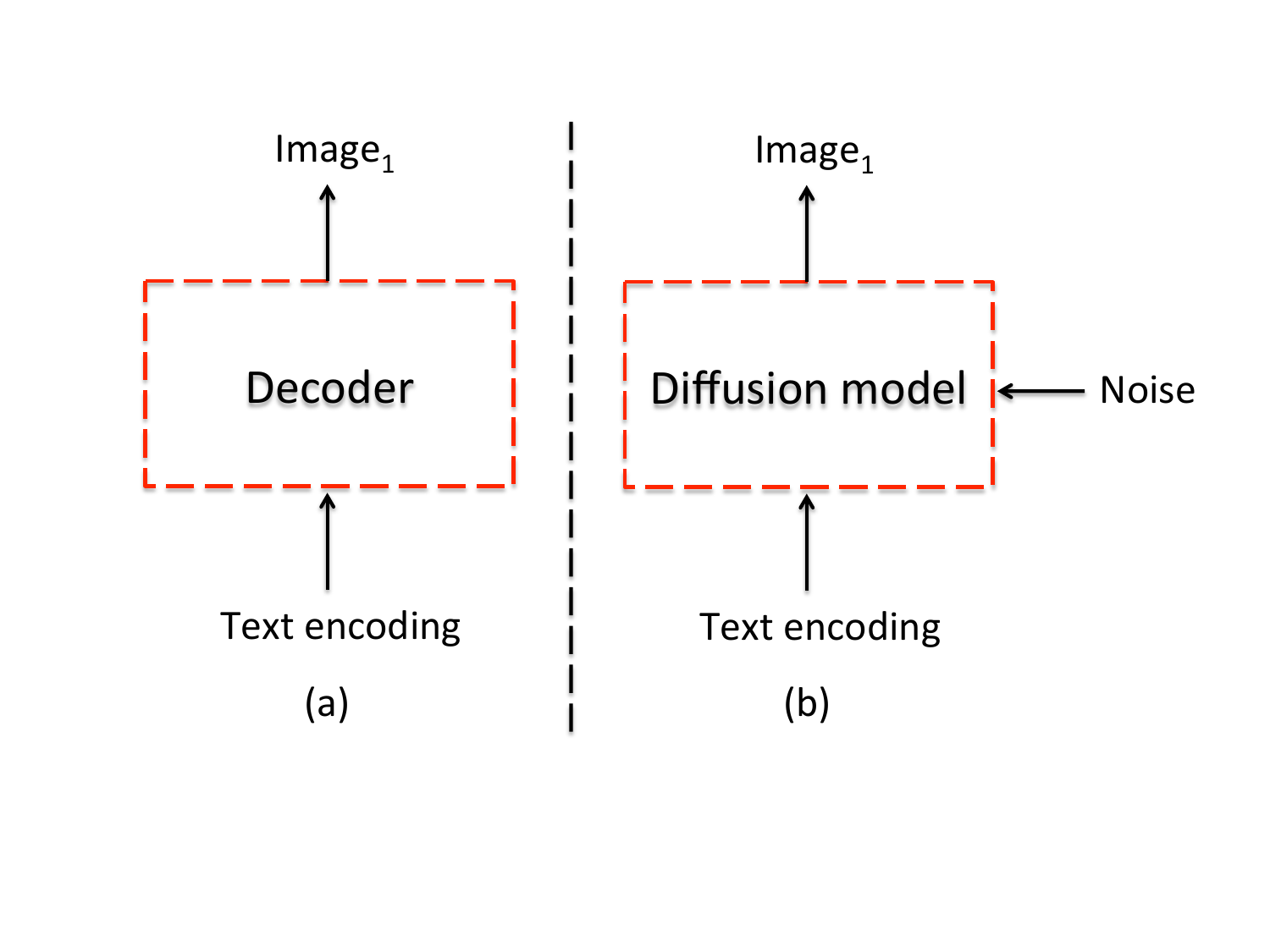}
\end{tabular}
\caption{The figure shows how to replace the Module in Fig 1. (a) Decoder; (b) Conditional diffusion model}
\label{Fig:module}
\end{figure}

To generate the video $Y=\{y_1, y_2,..., y_m \}$ as the sequence of frames, we need to take the similar approach as we encode the tokens in the transformer framework in Fig. \ref{Fig:rlcomp}. Since texts and images are from different domains, we need to map them into the same hidden space, which are further input to transformer. In Fig. \ref{Fig:rlcomp}, we have a ``Module" marked by red dash line to generate image. For example, we can either use decoder or conditional diffusion model from Fig. \ref{Fig:module} to do this job. As for conditional diffusion model, we can use the pretrained model and then need another deep nets to capture the motion information from the video. Although conditional diffusion model \cite{NicholDRSMMSC22,Saharia2022photorealistic} can generate high resolution image, it is time consuming and computing intensive. In this paper, we use decoder in Fig. \ref{Fig:module}(a) and train the whole model in an end-to-end fashion.

In other words, we have an encoder function $e: y \rightarrow h$ and a decoder $d: h \rightarrow y$. And we also require the generated $\hat{Y}$ matches the ground truth $Y$. We can minimize the following square error:
\begin{align}\label{eq:cerl}
loss = J(\theta)  =  \sum_{i=1}^m |y_i - \hat{y}_i|^2
\end{align}
where $\hat{y}_i = d( h_{<i})$ and $h_{<i}$ is the last hidden output at location $i$ from Transformer. The square error loss above on images is similar to the cross entropy in language models.

Another assumption that we make is that the image signal may become weak in the long sequential video. To enhance the signal, we take a similar approach from diffusion model. Thus, we use U-Net \cite{RonnebergerFB15} to construct images from its noised versions. 
The process is as follows:
\begin{enumerate}
\item create the noised data $\bar{y}_i = (1-\beta_i) y_i + \beta_i \epsilon$, where $\beta_i$ is the noise level coefficient

\item encode $h_i = e(\bar{y}_i) $ using the $down$ module from U-Net($down, up$), where we use the down module as our encoder

\item predict $h_{i+1}$ using the transformer 

\item decode the output $\hat{y}_{i+1} = d(h_{i+1})$ and reconstruct the $\hat{\bar{y}}_i = up(down(\bar{y}_i))$ with the $up$ module from U-Net($down, up$)

\item update the model parameters by minimizing the loss equation \ref{eq:loss} below
\end{enumerate}
\begin{align}\label{eq:loss}
loss & = L(\theta) + \alpha J(\theta)  =  \sum_{i=1}^n L_i(\theta) + \lambda J(\theta)  \nonumber \\
        &  = - \sum_{i=1}^n  \log p(x_i) + \lambda_1 \sum_{j=1}^m |y_j - \hat{y}_j |^2  +\lambda_2 \sum_{j=1}^m |y_j - \hat{\bar{y}}_j |^2 - \alpha| \hat{y}_{s\in [0,m]} - \hat{y}_{t \in [0,m]} |
\end{align}
where the first term is the cross entropy loss from the text part, the second is the reconstructing loss from the video part, and the last loss is to avoid the concentration in $\hat{y}$. And $\lambda = \{\lambda_1, \lambda_2 \}$ and $\alpha$ are the weights to balance these terms. As for the reconstruction loss, we consider both losses from reconstruction decoder ($\hat{y}$) and U-Net ($\hat{\bar{y}}_j$). As for the last loss $| \hat{y}_s - \hat{y}_t |$, we want to promote the motion between any two frames $(s,t)$. In the implementation, we randomly sample two frames $s,t \in[0, m)$, and we want the difference between these two frames as large as possible.


\section{Experimental results}

\begin{figure*}[h!]\center
\begin{tabular}{cccccc}
\includegraphics[trim=13mm 8.0mm 20mm 13.5mm, clip, width=2.0cm]{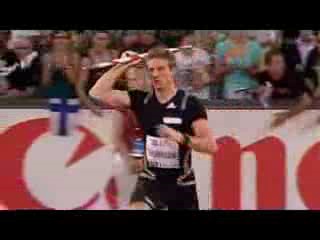} &
\includegraphics[trim=13mm 8.0mm 20mm 13.5mm, clip, width=2.0cm]{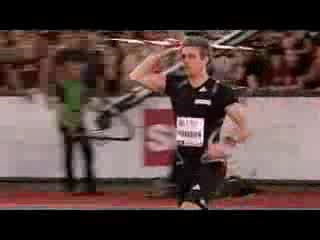} &
\includegraphics[trim=13mm 8.0mm 20mm 13.5mm, clip, width=2.0cm]{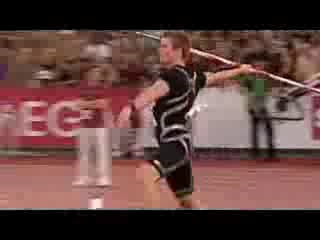} & 
\includegraphics[trim=13mm 8.0mm 20mm 13.5mm, clip, width=2.0cm]{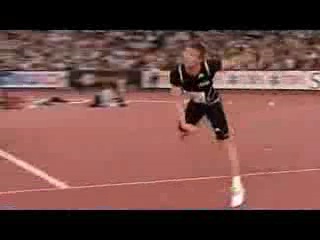} &
\includegraphics[trim=13mm 8.0mm 20mm 13.5mm, clip, width=2.0cm]{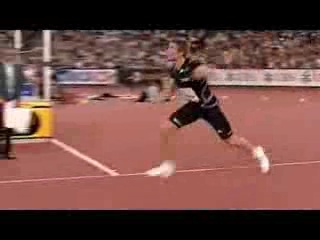} & 
\includegraphics[trim=13mm 8.0mm 20mm 13.5mm, clip, width=2.0cm]{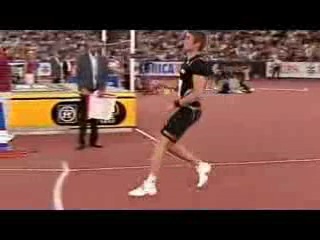} \\
\includegraphics[trim=13mm 8.0mm 20mm 13.5mm, clip, width=2.0cm]{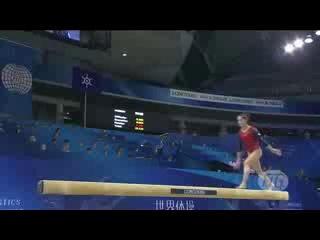} &
\includegraphics[trim=13mm 8.0mm 20mm 13.5mm, clip, width=2.0cm]{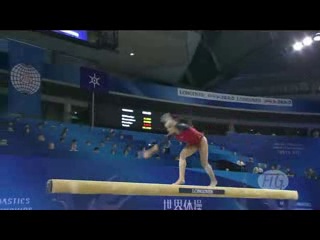} & 
\includegraphics[trim=13mm 8.0mm 20mm 13.5mm, clip, width=2.0cm]{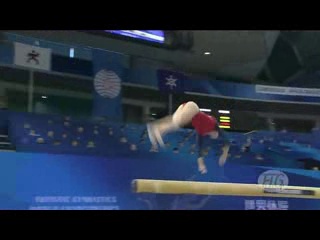} &
\includegraphics[trim=13mm 8.0mm 20mm 13.5mm, clip, width=2.0cm]{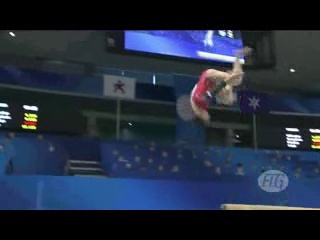} & 
\includegraphics[trim=13mm 8.0mm 20mm 13.5mm, clip, width=2.0cm]{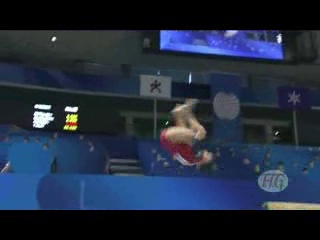}  &
\includegraphics[trim=13mm 8.0mm 20mm 13.5mm, clip, width=2.0cm]{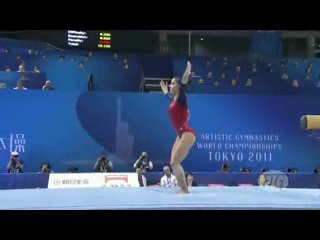} \\
\includegraphics[trim=13mm 8.0mm 20mm 13.5mm, clip, width=2.0cm]{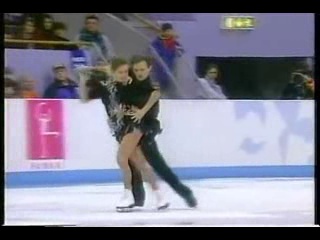} &
\includegraphics[trim=13mm 8.0mm 20mm 13.5mm, clip, width=2.0cm]{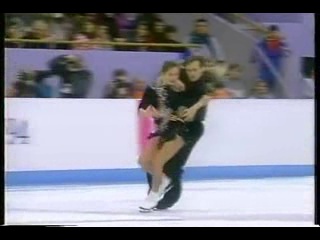} & 
\includegraphics[trim=13mm 8.0mm 20mm 13.5mm, clip, width=2.0cm]{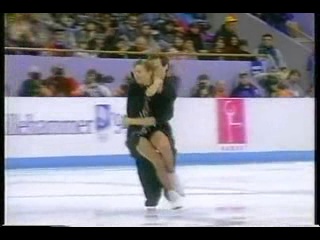} & 
\includegraphics[trim=13mm 8.0mm 20mm 13.5mm, clip, width=2.0cm]{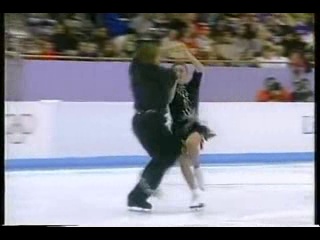}  &
\includegraphics[trim=13mm 8.0mm 20mm 13.5mm, clip, width=2.0cm]{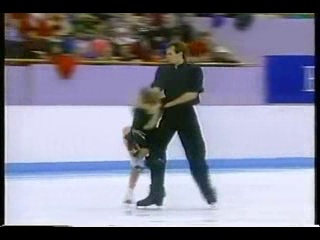} &
\includegraphics[trim=13mm 8.0mm 20mm 13.5mm, clip, width=2.0cm]{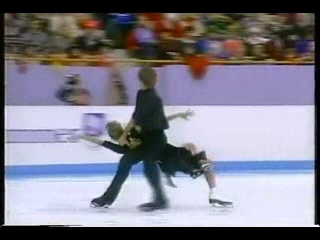} 
\end{tabular}
\caption{The sampled frames from 3 videos arranged in rows. The corresponding descriptions for each video is listed as: (1) A man with black clothes is throwing javelin with face forward from right to left; (2) A female gymnast performing front flip on balance beam from right to left; (3) A man and a woman with black clothes are doing ice dancing.}
\label{Fig:traingset}
\end{figure*}

We used the smallest version of GPT-2 with 124M parameters and U-Net , and tested our approach on UCF101 dataset \footnote{https://www.crcv.ucf.edu/data/UCF101.php}. There are total 101 types of actions in UCF101 dataset, and we select 60 actions and sample about 1 to 5 videos for each action. We label each video with a text description and then resize the video into $72\times72$ pixels to construct the final training dataset. In Fig. \ref{Fig:traingset}, we show the sampled (text, video) pair, where $X$ is the text and $Y$ is the video that we want to generate.

Because of limited computation resources, we resize the image to $32\times32$ and use a simple U-Net to encode it before transformer. The decoder is a simple linear layer with $tanh$ activation function, then it is resized to  $72\times72$ to reconstruct the output image. We also tried a four layer decoder, with each layer of cov2d, reshape and relu module, but it did not gain much in the reconstructed images. In the experiment, we found that layer normalization does not help to generate images, so we do not use it in the decoder. 
U-Net parameters: the base dim is 16, with dimensional multiples as (1, 2, 4, 8) and the num of resnets blocks as 2. 
We set  $\lambda =1$ for the first order image difference and $\lambda =5$ for the second order image difference. And we set $\alpha = 10$ to promote motion between any reconstructed image pair.

%

\begin{figure*}[h!]\center
\begin{tabular}{cccccc}
\includegraphics[trim=0mm 0mm 0mm 0mm, clip, width=2.0cm]{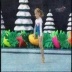} &
\includegraphics[trim=0mm 0mm 0mm 0mm, clip, width=2.0cm]{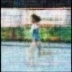} & 
\includegraphics[trim=0mm 0mm 0mm 0mm, clip, width=2.0cm]{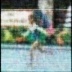} & 
\includegraphics[trim=0mm 0mm 0mm 0mm, clip, width=2.0cm]{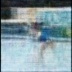}  &
\includegraphics[trim=0mm 0mm 0mm 0mm, clip, width=2.0cm]{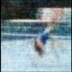} &
\includegraphics[trim=0mm 0mm 0mm 0mm, clip, width=2.0cm]{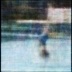} 
\end{tabular}
\caption{A girl with white clothes is performing the floor gymnastics from right to left.}
\label{Fig:result1}
\end{figure*}

We test our model with text prompts and show sampled images from generated videos in Fig. \ref{Fig:result1}. It shows that the reconstructed image resolution is not good enough. The reasons can be (1) the training set is small, and the image resolution is low; (2) the model structure is simple; (3) the decoder module is not good because the generated images become blurry as we increase the sequence length.
 
\begin{figure*}[h!]\center
\begin{tabular}{cccccc}
\includegraphics[trim=0mm 0.0mm 0mm 0.mm, clip, width=2.0cm]{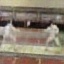} &
\includegraphics[trim=0mm 0.0mm 0mm 0.mm, clip, width=2.0cm]{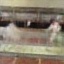} & 
\includegraphics[trim=0mm 0.0mm 0mm 0.mm, clip, width=2.0cm]{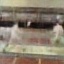} & 
\includegraphics[trim=0mm 0.0mm 0mm 0.mm, clip, width=2.0cm]{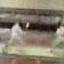}  &
\includegraphics[trim=0mm 0.0mm 0mm 0.mm, clip, width=2.0cm]{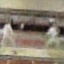} &
\includegraphics[trim=0mm 0.0mm 0mm 0.mm, clip, width=2.0cm]{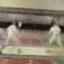} 
\end{tabular}
\caption{Two men wearing fencing suit practicing with sword against each other}
\label{Fig:result2}
\end{figure*}

\begin{figure*}[h!]\center
\begin{tabular}{cccccc}
\includegraphics[trim=0mm 0mm 0mm 0mm, clip, width=2.0cm]{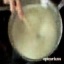} &
\includegraphics[trim=0mm 0mm 0mm 0mm, clip, width=2.0cm]{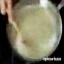} & 
\includegraphics[trim=0mm 0mm 0mm 0mm, clip, width=2.0cm]{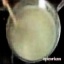} & 
\includegraphics[trim=0mm 0mm 0mm 0mm, clip, width=2.0cm]{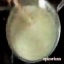}  &
\includegraphics[trim=0mm 0mm 0mm 0mm, clip, width=2.0cm]{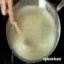} &
\includegraphics[trim=0mm 0mm 0mm 0mm, clip, width=2.0cm]{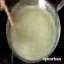} 
\end{tabular}
\caption{Stir the flour and water combination into source on the pan.}
\label{Fig:result3}
\end{figure*}

\begin{figure*}[h!]\center
\begin{tabular}{cccccc}
\includegraphics[trim=0mm 0mm 0mm 0mm, clip, width=2.0cm]{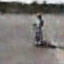} &
\includegraphics[trim=0mm 0mm 0mm 0mm, clip, width=2.0cm]{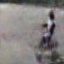} & 
\includegraphics[trim=0mm 0mm 0mm 0mm, clip, width=2.0cm]{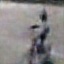} & 
\includegraphics[trim=0mm 0mm 0mm 0mm, clip, width=2.0cm]{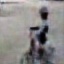}  &
\includegraphics[trim=0mm 0mm 0mm 0mm, clip, width=2.0cm]{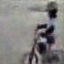} &
\includegraphics[trim=0mm 0mm 0mm 0mm, clip, width=2.0cm]{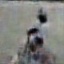} 
\end{tabular}
\caption{A boy in white t-shirt is biking with helmet.}
\label{Fig:result4}
\end{figure*}


\section{Conclusion}
We present a simple text to video model via transformer. Specifically, we combine both Transformer and U-Net to handle sequential text and long video datasets, and train the model in a end-to-end manner in order to generate videos from text prompt. The limited training dataset such as (text, video) pair is still a challenge to train a good model at this moment. In addition, we need to caption motion which should be object insensitive. In the next stage, we will try to improve quality with more datasets and complex models, such as conditional diffusion model to generate images from text, and transformer to capture motion in an object-insensitive manner.
%
%


\bibliographystyle{unsrt}
\bibliography{llm2022}

\end{document}